\documentstyle{llncs}
\input epsf
\begin{document}

\title{Exploring the Applications of Faster R-CNN and Single-Shot Multi-box Detection in a Smart Nursery Domain}

\author{Somnuk Phon-Amnuaisuk$^{1,2}$, Ken T. Murata$^3$, Praphan Pavarangkoon$^3$, Kazunori Yamamoto$^3$, Takamichi Mizuhara$^4$}
\institute{$^1$Media Informatics Special Interest Group, CIE, Universiti Teknologi Brunei \\
$^2$School of Computing and Information Technology, Universiti Teknologi Brunei \\
$^3$National Institute of Information and Communications Technology, Tokyo, Japan\\
$^4$CLEALINKTECHNOLOGY Co., Ltd., Kyoto, Japan\\
\email{somnuk.phonamnuaisuk@utb.edu.bn; ken.murata; praphan; kaz-y@nict.go.jp; mizuhara@clealink.jp}
}

\maketitle

\begin{abstract}
The ultimate goal of a baby detection task concerns detecting the presence of a baby and other objects in a sequence of 2D images, tracking them and understanding the semantic contents of the scene. 
Recent advances in deep learning and computer vision offer various powerful tools in general object detection and can be applied to a baby detection task. 
In this paper, the Faster Region-based Convolutional Neural Network and the Single-Shot Multi-Box Detection approaches are explored. They are the two state-of-the-art object detectors based on the region proposal tactic and the multi-box tactic.
The presence of a baby in the scene obtained from these detectors, tested using different pre-trained models, are discussed. This study is important since the behaviors of these detectors in a baby detection task using different pre-trained models are still not well understood. This exploratory study reveals many useful insights into the applications of these object detectors in the smart nursery domain.
\end{abstract}
\subsubsection*{Keywords}
Object detection; Faster Region-based Convolutional Neural Network (Faster R-CNN); Single Shot Multibox Detection (SSD); Smart nursery; 

\section{Introduction}
Smart nursery is a niche market that has been quickly growing in the past decades. This is attributed to the economic pressure in our modern lifestyle where both parents often opt to work to increase their income. Baby monitoring gadgets are emerged as parenting tools in the urban lifestyle. It provides a means for parents to monitor the well-being of their babies when they have to attend to other chores and cannot be present in the same physical space. The common functionalities of baby monitoring products are movement detection, breathing detection, remote visual and audio monitoring.

Contemporary visual and audio monitoring devices provide the means to transmit visual and audio information. Imagine if the monitoring gadget could report a baby's activities whether the baby is sitting, crawling, sleeping face up/down, etc., then this will be a very useful value-added alert functionality. A summary of the baby's activities could also give valuable information for the parents to evaluate the development of the babies. To date, this kind of application still does not exist. With recent advances in artificial intelligence, visual and audio processing, we attempt to explore this area, starting from the detection functionality. 

Object detection, localization, tracking and recognition are important prerequisites to the scenic understanding task which is still an open research problem. There are many open challenges from the following issues: variants introduced by transformations e.g., translation, scaling, and rotation; variants introduced by changes in lighting conditions, and colors; variants introduced by occlusion; and variants introduced by deformation in shape. These issues are the main challenges for researchers in computer vision community. 

With recent advances in convolutional neural networks \cite{simoyan15}, the last few years have seen many new robust techniques devised to handle the challenges mentioned above. Region-based Convolutional Neural Networks (R-CNN) \cite{girshick14}, Fast R-CNN \cite{girshick15}, Faster R-CNN \cite{ren15}, and Single-Shot Multi-box Detector (SSD) \cite{liu16} are examples of state-of-the-art visual object detection algorithms. In this paper, we explore the applications of the Faster R-CNN and the SSD both of which are the latest state-of-the-art detectors. Both techniques are employed to detect the presence of a baby from a video footage obtained from Youtube\footnote{Youtube standard license for fair use of public content}.

The rest of the materials in the paper are organized into the following sections; Section 2: Overview of the object detection task; Section 3: Exploratory study and Discussion; and Section 4: Conclusion.

\section{Overview of the Object Detection Task}
In order to detect whether an instance of a semantic object is in a 2D image, one could create a classification model trained with features extracted from positive and negative classes and use the trained model to classify a given image. This feature-based approach requires that the training images must be in the same nature as the testing images. Variations in position, color, lighting conditions, size, etc., greatly affect the performance of the model.

Object detection and tracking process are commonly fine-tuned to the problem at hand. This is usually dictated by the characteristics of the target. Detecting and tracking vehicles, require different parameter tuning than detecting pedestrians, persons face, or a baby, etc. Different targets require different detailed process due to the variations introduced by the environmental conditions e.g., target size, change in appearance. 

Feature-based techniques exploit the discriminative features of the objects derived from 2D pixel information in an image \cite{gonzalez01}. \emph{color}, and \emph{intensity} are the primitives where other structures could be derived from  e.g., \emph{edge}, \emph{contour}, \emph{color histogram}, and \emph{intensity gradient}, etc. Intensity gradient peaks are robust to illumination variations, they have a good repeatability and provide good descriptions of local appearances described as points. Detectors such as Kanade-Lucas-Tomasi (KLT) corner detectors \cite{tomasi94},  Scale Invariant Feature Transform (SIFT) \cite{lowe99} and Speeded Up Robust Features (SURF) \cite{bay08} successfully exploit this aspect of the feature-based technique.

Features such as points and gradient may be used to describe local structure. Composite structures such as Haar cascade and Histogram of Gradient (HOG) are successful descriptors that effectively describe the appearances of objects. Haar feature effectively abstracts human face through the abstraction of high/low intensity pattern of the human face e.g., forehead is the brighter region as compared to the eyes; HOG feature effectively abstracts a human figure using histogram of gradient, etc.

To increase the flexibility of the traditional feature-based approach, one could create a model based on a cropped version of the semantic object in the image and search for the object in a test image by scanning through it using different window sizes. This is plausible but the flexibility to deal with the variant of scale, rotation, deformation etc., comes with computing cost since the process requires many computation passes to handle each different variation.

\subsection{Faster Region-based Convolutional Neural Network}
Although it is plausible to find an instance of an object in a 2D image by sliding different shapes and sizes of search windows over the entire image, it is expensive to exhaustively scan through the image this way. In \cite{uijlings13}, the authors propose a concept known as \emph{selective search} (SS) where the region of interests in the image is proposed based on the similarity of local appearance. 
\begin{figure}[!ht]
\begin{center}\leavevmode
\epsfxsize=10cm
\epsfbox{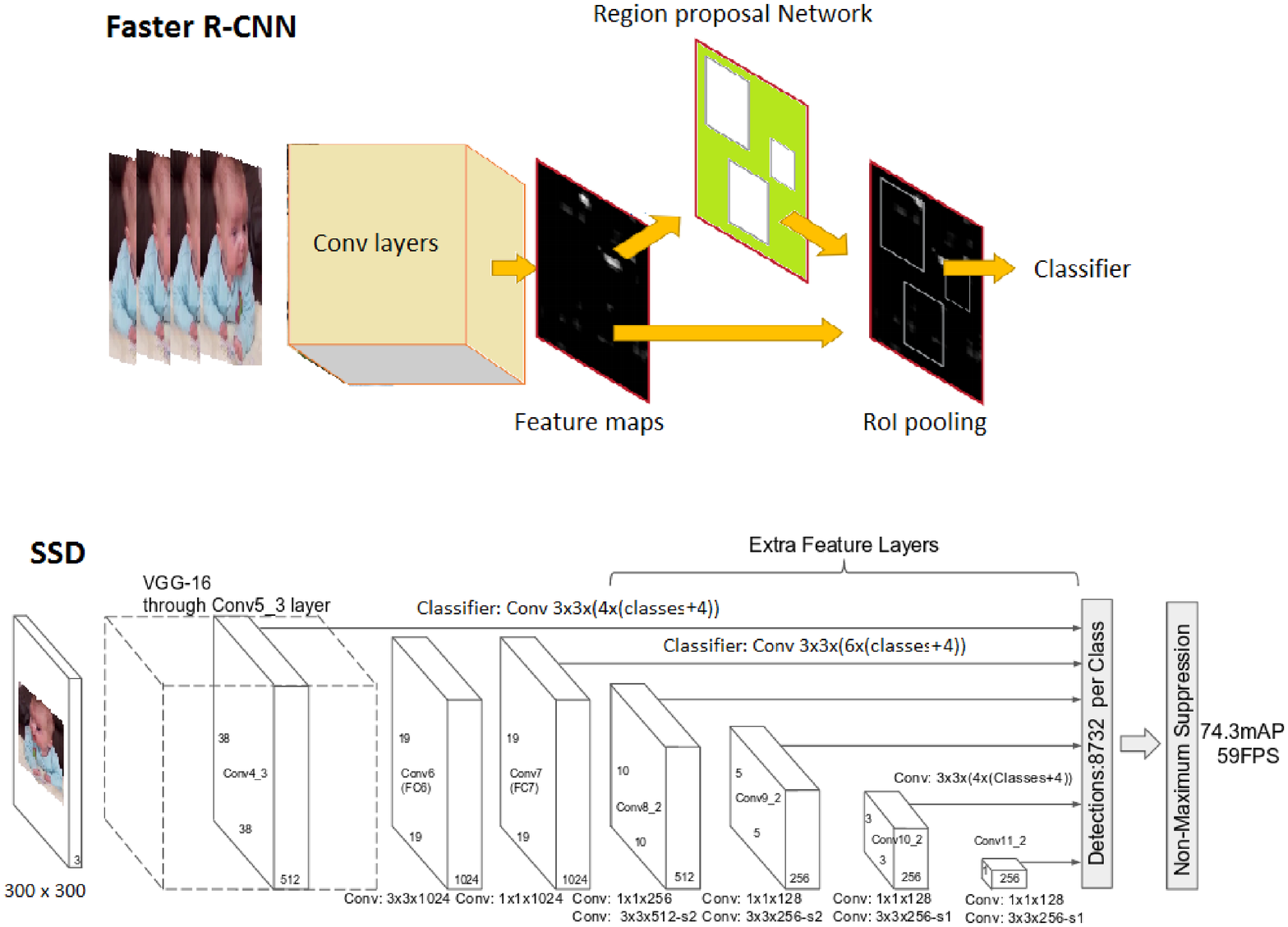}
\end{center}
\caption{Top pane: the Faster R-CNN approach (figure is adapted from \cite{ren15}). Bottom pane: the Single Shot Multibox Detection approach (figure is adapted from \cite{liu16}).}
\label{approach}
\end{figure}

Instead of performing exhaustive sliding window search, R-CNN and Fast R-CNN employ SS to provide over 2,000 proposed regions (per image) for its object detection search. Two thousand regions may seem a lot but this is just a fraction of the amount required by the exhaustive sliding window search alternative. The combination of SS and CNN allows instances of interested objects to be allocated in almost real-time (R-CNN can process one image in fifty seconds \cite{girshick14} and Fast R-CNN can process one image in two seconds). 

Faster R-CNN is an improvement of R-CNN and Fast R-CNN. Both R-CNN and Fast R-CNN have external region proposal process which requires a considerable amount of processing time. Faster R-CNN further improves the processing pipeline by embedding the region proposal network after the CNN (see Figure \ref{approach}). The region proposal process takes less than 10 ms to process the region proposal task and the Faster R-CNN offers the processing speed up to five frames per second which is good enough for some real life applications. 

In this study, the experiments with Faster R-CNN are carried out using the following pre-trained detection models obtained from Google \emph{Tensorflow}\footnote{https://github.com/tensorflow/models/blob/master/research/object\_detection/}: \emph{Inception} and \emph{ResNet}. These models are trained on the Microsoft \emph{Common Object Content} (COCO) dataset\footnote{http://cocodataset.org/}.

\subsection{Single-Shot Multi-Box Detection}
Leveraging on established deep convolutional neural network classifiers such as VGG network \cite{simoyan15}, the authors in \cite{liu16} propose \emph{Single-Shot Multi-Box Detection} (SSD) for an object detection task. SSD offers a novel object detection approach which separates itself from previous region-based approaches (R-CNN, Fast R-CNN and Faster R-CNN). SSD incorporates class prediction and bounding box prediction in a single process, hence, the detection speed of SSD is significantly faster than the \emph{Faster R-CNN} approach.  The SSD architecture consists of a base convolutional neural network\footnote{Base convolutional neural network refers to a CNN network without classification layers.} followed by the multi-box convolutional layers (see Figure \ref{approach}, bottom pane). The base VGG and multi-box layers predict (i) the presence of object class instances and (ii) their bounding box locations.  There are six prediction layers decreasing in feature-map size which allows the SSD to handle object detection in multi-scale. The total six layer feature maps with the sizes: $38^2, 19^2, 10^2, 5^2, 3^2$, and $1^2$, is implemented with the following numbers of default boxes: 4, 6, 6, 6, 4, and 4. Hence there are 8732 detections per class i.e., $38^2 \times 4 + 19^2 \times 6 + 10^2 \times 6 +  5^2 \times 6 +  3^2 \times 4 + 1^2 \times 4$ (see Figure \ref{approach}, bottom pane). Among these detections, many will be unlikely candidates with very small probability and they will be removed using the Non-Maximum Suppression (NMS).

Experiments with SSD are carried out using the base VGG network with the pre-trained \emph{PASCAL Visual Object Classes (VOC)} model\footnote{http://host.robots.ox.ac.uk/pascal/VOC/voc2007/} and the recent \emph{MobileNet} implementation \cite{howard17} from Google \emph{Tensorflow} with the pre-trained COCO dataset model. 

\begin{figure}[!ht]
\begin{center}\leavevmode
\epsfxsize=10cm
\epsfbox{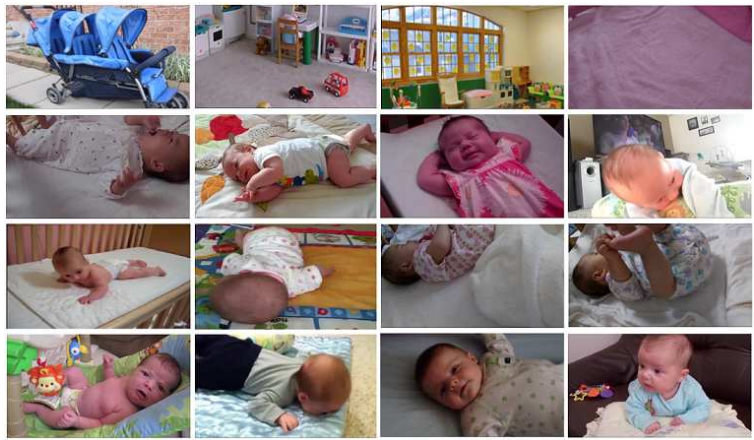}
\end{center}
\caption{Input to our object detection system is image frames (each frame should be at least 300x300 pixels$^2$). The top most row shows examples of negative input and the three bottom rows shows a typical input. The target object (baby) may be in various poses and orientations, and may be displayed full body or with occlusion. The environment may have different lighting conditions.}
\label{task}
\end{figure}
\section{Exploratory Study and Discussion}
Due to the non-existence of any standard baby monitoring benchmark dataset, we prepare our test data using in-house footage as well as by editing video clips of baby (babies are less than twelve months old) found on Youtube. These Youtube clips are utilized under the Youtube's standard fair use license. These clips are shot and posted by parents. Hence, these clips are shot with unprofessional setting, with different lighting environments, various camera angles and with camera movement. This imperfection provides reasonable variants for us to evaluate the usage of pre-trained detector in the actual environment. Figure \ref{task} provides examples of the test images used in this paper.

\subsection{Evaluation Criteria}
We are interested to know the behaviors of the pre-trained Faster R-CNN and the SSD detectors in detecting the presence of a baby in a video sequence. Since the Faster R-CNN and the SSD detectors are capable of detecting more than one class, in this setup, the evaluation is based on the correct detection and classification of the baby in the scene but not for other objects. In other words, a frame is classified as a true positive (TP) if there is a baby in the scene and the detector has correctly detected the baby; as a true negative (TN) if a baby is not in the scene and the detector does not output any detection of a baby; as a false positive (FP) if a baby is not in the scene but the detector wrongly reports the detection; and as a false negative (FN) if a baby is  in the scene but the detector does not output any detection of a baby, or wrongly labels the detected baby. Figure \ref{exp} displays examples of TP, TN, FP and FN cases.  We calculate the sensitivity, the specificity and the accuracy using the equations below:
\begin{figure}[!ht]
\begin{center}\leavevmode
\epsfxsize=10cm
\epsfbox{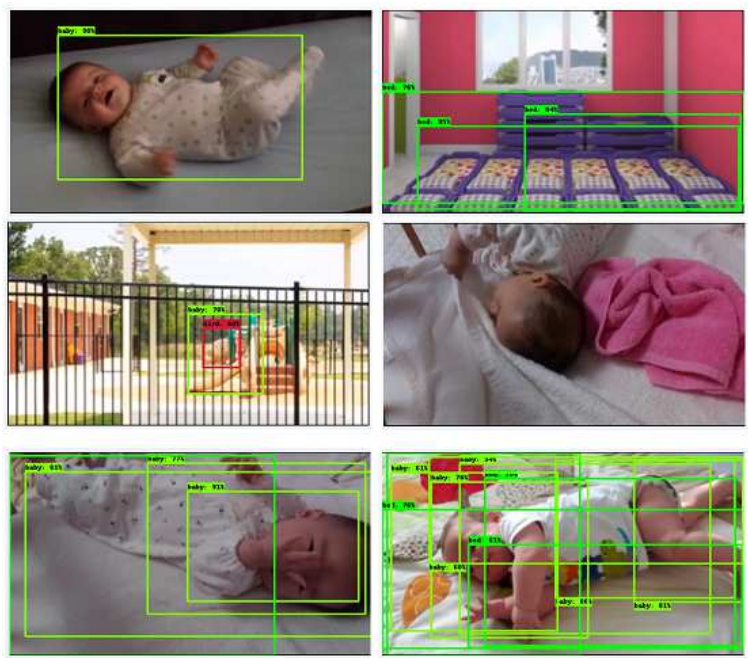}
\end{center}
\caption{Representative examples of (top row) true positive and true negative cases; (middle row) false positive and false negative cases. Bottom row: the detectors may suggest many bounding box for the same object. This is considered as a correct classification. This detection behavior is, however, not preferable since it complicates the precision of the localization task.}
\label{exp}
\end{figure}
\begin{small}
\begin{equation}
Sensitivity = \frac{\mbox{True positive}}{\mbox{True positive + False negative}}  \end{equation}
\begin{equation}
Specificity = \frac{\mbox{True negative}}{\mbox{True negtive + False positive}}  \end{equation}
\begin{equation}
Accuracy = \frac{\mbox{True positive + True negative}}{\mbox{True positive + True negtive + False positive + False negative}}  
\end{equation}
\end{small}

Each test clip is manually prepared using the content from the Youtube video clips. It is decided to have 100 positive examples and 50 negative examples in each test clip. The mp4 movies and image sequences of the test clips can be accessed from this url \texttt{https://sites.google.com/view/eventanalysis/home/datasets}

\subsection{Discussion}
We employ both Faster R-CNN and SSD to detect the presence of a baby. The detectors are chosen as they are the representative of the current state-of-the-art detectors. 
The results from this exploratory study are tabulated in Table \ref{results}. The results show a reasonable detection accuracy from all models\footnote{Pre-trained models: faster\_rcnn\_inception\_v2\_coco, faster\_rcnn\_ResNet101\_coco, ssd\_mobilenet\_v2\_coco and ssd300\_mAP\_77.43\_v2}. The accuracy of  Faster R-CNN in detecting the presence of baby appears to outperform the performance from SSD. However, from our observations (see Figure \ref{exp2}), SSD is faster and gives higher quality bounding box than Faster R-CNN.

\begin{table}
\begin{center}
\begin{small}
\caption{Summary of sensitivity, specificity and average accuracy of the Faster R-CNN and the SSD detectors. Microsoft COCO is the training dataset for all models except the SSD VGG model which uses the VOC dataset.}
\label{results}
\begin{tabular}{|c|cc|cc|cc|cc|cc|c|} \hline  \hline
 & \multicolumn{2}{c|}{Footage 1} & \multicolumn{2}{c|}{Footage 2} & \multicolumn{2}{c|}{Footage 3} & \multicolumn{2}{c|}{Footage 4} & \multicolumn{2}{c|}{Footage 5} & Acc \\
{\bf Model} & Sens & Spec & Sens& Spec & Sens & Spec & Sens & Spec & Sens & Spec & (\%)\\  \hline 
{\bf Faster R-CNN}  & & & & & & & & & & &\\
Inception model (COCO) & 0.98& 1.00& 0.94 & 1.00& 1.00 & 0.84& 0.96& 0.96& 0.98& 0.80 & {\bf 95.5}\\
ResNet model (COCO) & 0.98& 1.00& 0.96 & 1.00 &1.00 & 0.98& 0.99& 0.86& 0.98& 0.96 & {\bf 97.5}\\ \hline
{\bf SSD} & & & & & & & & & & &\\
MobileNet model (COCO) & 0.80& 1.00& 0.70 & 0.80&0.84 & 1.00& 0.87& 1.00& 0.79& 1.00 & {\bf 85.3}\\
VGG model (VOC) & 0.88& 1.00& 0.70 & 0.56&0.99 & 0.92& 0.90& 0.86& 0.86& 0.92 & {\bf 86.1}\\ \hline
\end{tabular}
\end{small}
\end{center}
\end{table}

Accurate detection, localization and tracking enhance the performance of the recognition system which in turn enable accurate scenic understanding.
This work investigates the performance of the two state-of-the-art detectors using various pre-trained CNN architectures: Inception model, ResNet model, MobileNet model and VGG model; with two detection approaches: Faster R-CNN and SSD.
Figure \ref{exp2} shows the representative detection examples from these models. We would like to highlight the following points to the readers:
\begin{figure}[!ht]
\begin{center}\leavevmode
\epsfxsize=12cm
\epsfbox{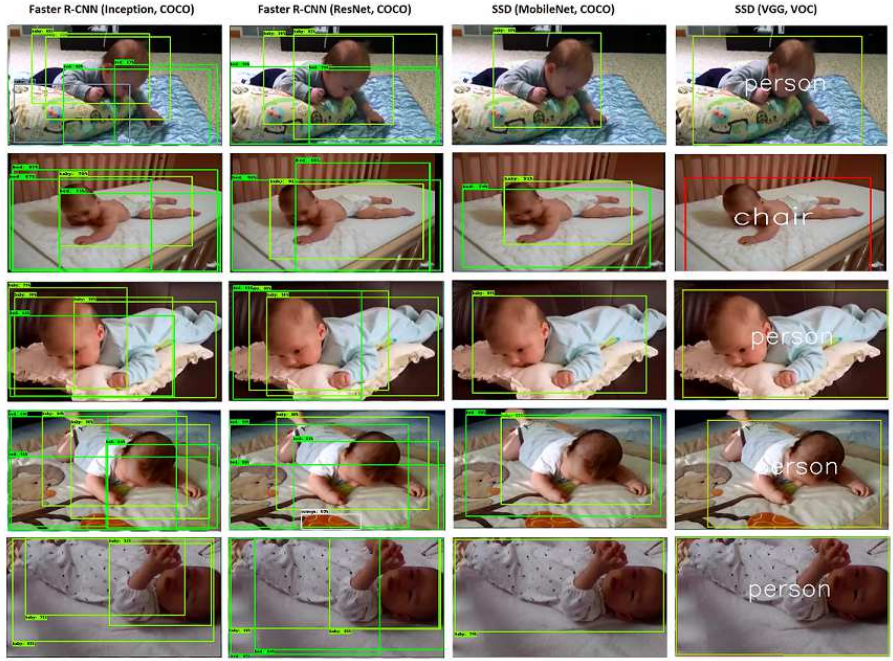}
\end{center}
\caption{Comparing behaviors of Faster R-CNN (columns one and two) and SSD (columns three and four), it is apparent that the Faster R-CNN predicts more bounding boxes of an object more than the SDD does. This is due to the non-maximum supression mechanism in SSD while the Faster R-CNN relies on the region proposal mechanism and does not provide the combining mechanism in the final stage.}
\label{exp2}
\end{figure}

\subsubsection{Quality of the bounding box prediction :} 
SSD appears to provide appropriate bounding boxes for its predictions while Faster R-CNN appears to offer many bounding boxes overlapping the interested regions (see Figure \ref{exp2}). Many bounding boxes are not desired since this will complicate the localization task.

\subsubsection{Quality of the class detection :}
Faster R-CNN provides a better baby detection accuracy (the best is 97.5\% with our dataset) while the accuracy from the SSD is only 86.1\%. This level of accuracy could be enhanced with various tactics such as (i) spatial and temporal continuity constraints on the detected baby; and (ii) ensemble learning where prediction from various models may be combined. These are important future work areas.

\subsubsection{The gap between detection and activity recognition :} 
Detection only detects an instance of a class but does not provide any distinction among instances of the same class. If we are interested in a counting task, e.g., counting the number of babies occupying a room, accurate detection is enough for the task. However, if we would like to collect their detailed profiles such as their activities, then the \emph{ID-tracking} and \emph{activity recognition} functionalities must be appended to the pipeline. ID-tracking enables each object to be tracked anonymously. Activity recognition provides detailed activity labels to the tracked objects. Common activity recognition techniques are based on motion descriptors derived from 2D image sequences such as \emph{Histograms of Oriented Optical Flow} (HOOF) \cite{chaudhry09}. 
Improvement of speed for the detection-recognition pipeline is also an important future work area.

\subsubsection{The gap between detection and scenic understanding :} 
In order to provide an appropriate summary of the visual content, accurate object detection alone is also not enough since relationships between detected components play a crucial role in scenic understanding. To date, this is still an open research area. From our analysis, we categorize the approaches toward bridging the gap between object detection and scenic understanding into two main categories (i) a knowledge-based framework and (ii) an encoder-decoder framework.

The knowledge-based framework attempts to explicitly construct objects and their relationships in the scene. For example, in a babywatching task, if a baby, a teddy bear and small Lego pieces are detected in the scene, then relationships between these objects can be represented as stereotyped situations \cite{minsky75} where various inferences may be drawn from them.  External knowledge is commonly injected into the knowledge system. This provides an effective inference process but  requires a hand-crafted knowledge encoding which is a well-known bottleneck in the knowledge engineering task.

The encoder-decoder framework is leveraged from recent advances in deep learning. A deep CNN is a common encoder and a long short-term memory (LSTM) network \cite{hochreiter97} is a typical decoder. An image frame is encoded into a feature vector  which can be decoded into appropriate natural language descriptions of the given frame \cite{vinyals16}. The encoder-decoder framework does not explicitly encode knowledge in the fashion employed in the knowledge-based approach. Hence, further processing steps are required to transform an image caption of each frame into a coherent scenic understanding.

\section{Conclusion}
The goal of object detection is to detect an instance of semantic objects. The approach has been constantly evolving in the past four decades. In the early days, the process was considered successful if the main object in the scene had been correctly identified. Later, localization has become a priority and it is normally provided as a bounding box over the object. With the progress in deep CNN \cite{liu16,szegedy15,krizhevsky12,sermanetoverfeat}, accurate detection and localization have been witnessed.

In this paper, we carried out experiments in the babywatching domain using state-of-the-art detectors. The results obtained from these detectors based on the recent pre-trained models from the Tensorflow detection model zoo have been carefully analyzed and discussed.  This reveals new insights and many potential future research in this domain that could bridge the gap between object detection and scenic understanding.

\begin{small}
\subsubsection*{Acknowledgments}
We would like to thank the Centre for Innovative Engineering (Universiti Teknologi Brunei), NICT (Japan) and CLEALINKTECHNOLOGY (Japan) for their supports given to this work. 
\end{small}

%
%

%
\end{document}